\documentclass[10pt, conference, compsocconf]{IEEEtran}
\usepackage{cite}
\usepackage{graphicx}
\usepackage{algorithm}
\usepackage{algorithmic}
\usepackage{amsmath}
\usepackage{multirow}
\usepackage{mathtools}
\usepackage[table]{xcolor}
\usepackage{multicol}
\usepackage{bbm}
\usepackage{url}
\usepackage{soul}
\usepackage{subcaption}

\newcommand\norm[1]{\lVert#1\rVert}

\ifCLASSINFOpdf
\else
\fi
\begin{document}
%
\title{Supervised Contrastive Learning for Detecting Anomalous Driving Behaviours\\from Multimodal Videos}


\author{\IEEEauthorblockN{Shehroz S. Khan\textsuperscript{1,2}, Ziting Shen\textsuperscript{2}, Haoying Sun\textsuperscript{2}, Ax Patel\textsuperscript{2}, Ali Abedi\textsuperscript{1}}
\IEEEauthorblockA{\textsuperscript{1}KITE – Toronto Rehabilitation Institute, University Health Network, Toronto, Canada\\
\textsuperscript{2}Institute of Biomedical Engineering, University of Toronto, Toronto, Canada\\
\{shehroz.khan, ali.abedi\}@uhn.ca, \{ziting.shen, haoying.sun, ax.patel\}@mail.utoronto.ca}


}


%


\maketitle

\begin{abstract}
Distracted driving is one of the major reasons for vehicle accidents. Therefore, detecting distracted driving behaviours is of paramount importance to reduce the millions of deaths and injuries occurring worldwide. Distracted or anomalous driving behaviours are deviations from 'normal' driving that need to be identified correctly to alert the driver. However, these driving behaviours do not comprise one specific type of driving style and their distribution can be different during the training and test phases of a classifier. We formulate this problem as a supervised contrastive learning approach to learn a visual representation to detect normal, and seen and unseen anomalous driving behaviours. We made a change to the standard contrastive loss function to adjust the similarity of negative pairs to aid the optimization. Normally, in a (self) supervised contrastive framework, the projection head layers are omitted during the test phase as the encoding layers are considered to contain general visual representative information. However, we assert that for a video-based supervised contrastive learning task, including a projection head can be beneficial. We showed our results on a driver anomaly detection dataset that contains $783$ minutes of video recordings of normal and anomalous driving behaviours of $31$ drivers from various top and front cameras (both depth and infrared). We also performed an extra step of fine tuning the labels in this dataset. Out of $9$ video modalities combinations, our proposed contrastive approach improved the ROC AUC on $6$ in comparison to the baseline models (from $4.23\%$ to $8.91\%$ for different modalities). We performed statistical tests that showed evidence that our proposed method performs better than the baseline contrastive learning setup. Finally, the results showed that the fusion of depth and infrared modalities from top and front view achieved the best AUC ROC of $0.9738$ and AUC PR of $0.9772$.


\end{abstract}

\begin{IEEEkeywords}
driving behaviours; anomaly detection; supervised contrastive learning; video anomaly detection

\end{IEEEkeywords}

%
\IEEEpeerreviewmaketitle

\section{Introduction}
\noindent According to the World Health Organization, approximately $1.3$ million people die each year due to road traffic accidents \cite{pietrasik_2021}. Distracted driving is one of the key risk factors, along with speeding, use of substances, and unsafe vehicles \cite{pietrasik_2021}. In many jurisdictions, such as in Ontario (Canada), distracted driving laws prohibit drivers to use phones or other hand-held wireless devices, to view display screens unrelated to driving, or to manually program GPS devices during driving \cite{government_of_ontario_2021}. In reality, distracted driving extends from such legal definitions to various day-to-day actions, such as drinking, eating, talking with passengers, combing hair, taking hands off the wheel and adjusting radio while driving. These actions increase the risk of distracted driving related collisions. 

Driver monitoring systems target to identify distracted or dangerous driving behaviours that could cause potential traffic accidents. It can assess a driver’s alertness, attentiveness, and focus and generates warning signals if distracted driving behaviour is observed. However, distracted driving does not comprise of a unique set of behaviours and could result due to negligence, carelessness, fatigue, or other unknown reasons. Most of the time a driver may be driving safely, yet a minor distraction could be fatal. Therefore, due to the diversity and less occurrence of these distracted behaviours, we term them as 'anomalous' driving behaviours. This should not be confused with traditional 'anomaly detection' problems where the anomalous events are not available during training phase. In our problem formulation, some annotated anomalous driving behaviours are available during training phase. From a machine learning perspective, this represents an imbalanced class classification problem with lots of normal or safe driving data and few anomalous driving samples. Moreover, the distribution of anomalous driving behaviours could be different during training and test phases due to variations in people exhibiting different types of anomalous driving behaviours. Hence, new anomalous driving behaviours may occur during testing that were not observed in the training phase.

For handling imbalanced classes (in videos), supervised 3D Convolutional Neural Network (3DCNN) or its variants with weighted classes could be the default choice \cite{li2022vbnet,kopuklu2020drivermhg}. However, more recently contrastive learning (CL) has shown state-of-the-art results in both self-supervised \cite{chen2020simple} and supervised classification problems \cite{khosla2020supervised}. CL \cite{chen2020simple} enables to learn effective visual representations by contrasting
positive and negative pairs;
thus, enabling them to work in imbalanced dataset situations. CL architecture is generally realized through an encoder (2D or 3D CNN or its variants), followed by projection head (feed forward layers). Supervised CL \cite{khosla2020supervised,Xu_2022_WACV,peeters2022supervised,liang2021enhancing,bommes2021anomaly} leverages the class labels present in the data to achieve state-of-the-art results in classification problems. In the denominator of a typical CL loss function, the exponential of dot product of negative pairs is summed. If the labelled negative pairs are closer to the positive pairs,
due to negative samples being near to positive samples (e.g.,. mislabeling), then this can lead to difficulty in the overall optimization of the loss function. This problem could be mitigated by scaling the sum term such that the summed similarity of negative pairs remain far to align with the loss function. To circumvent this potential problem, we proposed to average the sum of negative pairs instead.

The projection head is generally discarded in self-supervised \cite{chen2020simple} and supervised CL approaches \cite{khosla2020supervised} with the assumption that the representation learned at the encoder stage captures a generic visual representation.
According to Bommes et al. \cite{bommes2021anomaly}, it is possible to detect anomalies in \textit{images} by maintaining the projection head.
We assert that for the \textit{video-based} supervised classification task, the visual representation learned at the projection head stage can be more helpful for the (imbalanced) classification task. 

We validated the above mentioned modifications on a driver anomaly detection (DAD) dataset containing nine combinations of four data modalities -- top and front fitted depth and Infrared Radiation (IR) cameras inside a driver simulator \cite{kopuklu2021driver}, 
comprising of a training set ($25$ participants, $650$ minutes) and test set ($6$ participants, $133$ minutes) with different normal and anomalous driving behaviours captured through these cameras. The majority of the anomalous behaviours in the test set are not present in the training set. 
Our two-fold contribution is, (i) the proposed contrastive loss function and (ii) utilizing the entire network, including the projection head, in the test phase that improved results on six out of nine camera modalities by $4.23\%$ to $8.91\%$.

\section{Related Work}

Anomaly detection problems mostly refer to situations where the data for the anomalous class (or positive class) is not available during the training phase either due to its rarity, difficulty in collection, and health or safety hazards \cite{khan2014one}. However, in many situations, some labelled data may be available for the positive class, albeit with a large skewed data distribution. Such problems can be handled in a supervised machine learning setup using different strategies, such as weighted loss functions and data augmentation. In the context of video-based anomaly detection, there exist many approaches based on autoencoders \cite{nogas2020deepfall,vu2019robust} and adversarial learning \cite{khan2021spatio,cai2021appearance,mehta2021motion}. Supervised approaches for video classification are mostly based on 
3D-CNN \cite{ji20123d} and/or combined with sequential models \cite{wu2015modeling,girdhar2019video}. However, these models may be difficult to train on highly skewed data, especially in cases where the distribution of positive or anomalous classes differs across the training and test set.

Contrastive learning \cite{hadsell2006dimensionality} approaches have been evolving around the idea of contrasting positive pairs against negative pairs, maximizing the `similarity' among positive pairs while minimizing the `similarity' between positive-and-negative pairs. The contrastive loss equation is also closely connected to N-pair loss, triplet loss, as the latter being a special case of generalized contrastive loss where the number of positives and negatives are each one \cite{khosla2020supervised}.

Chen et al. \cite{chen2020simple} added a projection head consisting of several feed-forward layers in front of an encoder that improved the quality of learned representation. They showed that models with unsupervised pre-training with CL achieved better than their supervised counterparts. Khosla et al. \cite{khosla2020supervised} extended this idea for supervised CL to leverage labels present in the data. Their results achieved better results with Resnet-200 on the ImageNet dataset. They also showed that supervised CL outperformed cross-entropy on other datasets and two ResNet variants. This approach was shown to be robust to data corruptions, and stable to the choice of optimizers and data augmentations. CL has been used in anomaly detection with success. Zheng et al. \cite{zheng2021generative} applied self-supervised CL on graph neural networks for anomaly detection and outperformed state-of-the-art methods on several datasets.

Kopuklu et al. \cite{kopuklu2021driver} introduced the driver anomaly dataset containing $783$ minutes of overall video data collected from $31$ individuals, using top and down fitted depth and IR cameras. They applied a supervised CL approach (the neural network architecture was the same as Chen et al. \cite{chen2020simple}) that achieved better results than cross-entropy loss on detecting anomalous driving behaviours. The work presented in this paper presents modifications on Kopuklu et al. \cite{kopuklu2021driver} in terms of a new loss function and including projection head during test phase
to obtain improved results on six out of nine camera modalities and support them with statistical testing.

\section{Video-based Driver Anomaly Detection Datasets}
Anomalous driving behaviours involve various types of distraction, such as looking/moving away head from the road, 
setting the radio/GPS, texting or calling on a cell phone, to name a few. To capture different types of driving distractions, several driving datasets are available. 
CVRR-HANDS 3D \cite{ohn2013driver} and
DriverMHG \cite{kopuklu2020drivermhg} are hand focused datasets that are correlated with a driver's ability to drive. In addition, datasets such as DrivFace \cite{diaz2016reduced}, DriveAHead \cite{schwarz2017driveahead} and DD-Pose \cite{roth2019dd} provides information on drivers' faces for paying attention and head pose annotations of yaw, pitch and roll angles. Drive\&Act \cite{martin2019drive} provides distraction related anomalous driving data for $5$ near-IR cameras. 
There are many other driver monitoring datasets that provide face, head, hands, and body actions to detect normal and distracted driving behaviours, e.g., DMD \cite{Ortega_2020}, and Pandora \cite{borghi2017poseidon}. A major limitation of these datasets is that the normal and anomalous driving samples are present in both the train and test set; thus indirectly simplifying the complexity of the problem. Anomalous behaviours vary across people, the  distribution of actions available during training may not be available during testing. The generalization capabilities of these datasets in real-world driving situations remain unknown. 

\begin{table*}[t]
\centering
\begin{tabular}{|l|lll|} \hline
\textbf{Anomalous Actions in Training Set} & \multicolumn{3}{c|}{\textbf{Anomalous Actions in Test Set}} \\ \hline
Talking on the phone-left & Talking on the phone-left & Adjusting side mirror & Wearing glasses \\
Talking on the phone-right & Talking on the phone-right & Adjusting clothes & Taking off glasses \\
Messaging left & Messaging left & Adjusting glasses & Picking up something \\
Messaging right & Messaging right & Adjusting rear-view mirror & Wiping sweat \\
Talking with passengers & Talking with passengers & Adjusting sunroof & Touching face/hair \\
Reaching behind & Reaching behind & Wiping nose & Sneezing \\
Adjusting radio & Adjusting radio & Head dropping (dozing off) & Coughing \\
Drinking & Drinking & Eating & Reading \\ \hline
\end{tabular}
\caption{List of anomalous actions in the training and test sets. The last two columns in the test set contain actions not seen in the training set.}
\label{tab:actions_list}
\end{table*}

\begin{figure}

    \begin{subfigure}[b]{0.45\linewidth}
    \centering
    \includegraphics[width=\linewidth]{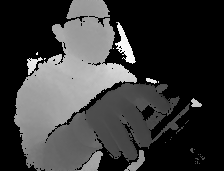}
    \caption{}
    \end{subfigure}
    \hfill
    \begin{subfigure}[b]{0.45\linewidth}
    \centering
    \includegraphics[width=\linewidth]{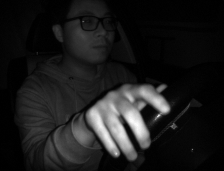}
    \caption{}
    \end{subfigure}
    \\[1ex]
    \begin{subfigure}[b]{0.45\linewidth}
    \centering
    \includegraphics[width=\linewidth]{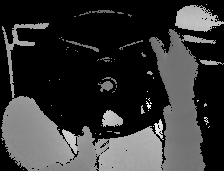}
    \caption{}
    \end{subfigure}
    \hfill
    \begin{subfigure}[b]{0.45\linewidth}
    \centering
    \includegraphics[width=\linewidth]{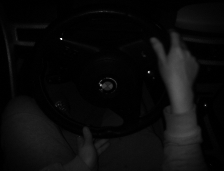}
    \caption{}
    \end{subfigure}
\caption{Normal driving behaviour frames captured from different views and modalities. (a) Front depth, (b) Front IR, (c) Top depth, (d) Top IR.}
\label{fig:dataset_pictures}
\end{figure}

\begin{figure}

    \begin{subfigure}[b]{0.45\linewidth}
    \centering
    \includegraphics[width=\linewidth]{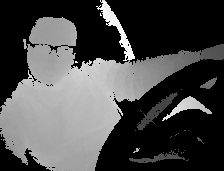}
    \caption{}
    \end{subfigure}
    \hfill
    \begin{subfigure}[b]{0.45\linewidth}
    \centering
    \includegraphics[width=\linewidth]{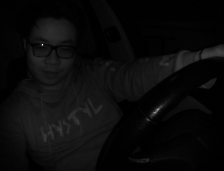}
    \caption{}
    \end{subfigure}
    \\[1ex]
    \begin{subfigure}[b]{0.45\linewidth}
    \centering
    \includegraphics[width=\linewidth]{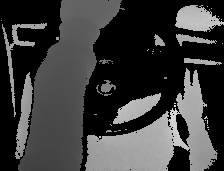}
    \caption{}
    \end{subfigure}
    \hfill
    \begin{subfigure}[b]{0.45\linewidth}
    \centering
    \includegraphics[width=\linewidth]{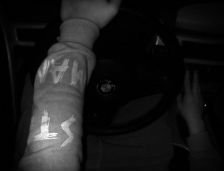}
    \caption{}
    \end{subfigure}
\caption{Anomalous driving behaviour frames (adjusting radio) captured from different views and modalities. (a) Front depth, (b) Front IR, (c) Top depth, (d) Top IR.}
\label{fig:anomalous_pictures_2}
\end{figure}
\subsection{DAD Dataset}
Kopuklu et al. \cite{kopuklu2021driver} presented the Driver Anomaly Detection (DAD) that overcomes the limitations of previous datasets. The DAD dataset is collected from a driver simulator containing a real BMW car cockpit, while the participants drive in a computer game that is projected in front of the car. Two Infineon CamBoard pico flexx cameras (depth and IR) were placed on top and in front of the driver. The front camera recorded the drivers' head, body and visible part of the hands, while the top camera recorded the driver's hands movements. Therefore, this dataset is both multi-view and multi-modal. The dataset includes a training set comprising of $650$ minutes of video data collected from $25$ participants, and a test set comprising of $133$ minutes of video data collected from $6$ participants. 
The 
video recordings in DAD dataset were synchronized with a resolution of $224 \times 171$ pixels and frame rate of $45$ fps. Within the training set, each participant performed $6$ normal driving and $8$ anomalous driving video recordings with the same time duration. In total, there are around $550$ minutes recording for normal driving and $100$ minutes recording of anomalous driving in the training set. Within the test set, each participant recorded $6$ video recordings that last around $3.5$ minutes. Anomalous actions were contained randomly within those test video clips. In total, the test set contains around $88$ minutes of normal driving and $45$ minutes of anomalous driving. In addition to samples from $8$ anomalous driving behaviours in the training set, the testing set contained $16$ new distracted actions (see Table \ref{tab:actions_list}). The anomalous actions in the DAD dataset include multiple criteria, such as head and body movements (reaching behind, talking with passengers), and hand interactions (messaging, drinking). By collecting a large amount of videos containing a wider amount of anomalous driving actions, the DAD dataset represents the real-life situation more comprehensively and is suitable for learning deep learning models. Figures \ref{fig:dataset_pictures} and \ref{fig:anomalous_pictures_2} show sample normal and anomalous driving actions captured from the depth and IR cameras from top and front views.




\section{Supervised Contrastive Learning Approach}

The network architecture for supervised CL framework consists of the following three components:

\begin{itemize}
    \item Encoding layers, $f_\theta$(.) -- A 3D-CNN that extracts vector representations ($h_i$) of input video clips ($x_i$), $h_i = f_\theta(x_i)$.
    
    \item Projection layers, $g_\beta$(.) -- comprises of fully connected layers to transform latent representation from the encoding layers to another latent representation, s.t. $v_i = g_\beta(h_i)$. The contrastive loss is defined on the \textit{l}2-normalized versions of this representations.
  
    
    \item Contrastive loss -- calculated from the summation of the similarities between vector representations of each normal video (anchor) and all other normal video clips (normal pairs) and the similarities between each anchor and all the anomalous video clips (anomalous pairs).

\end{itemize}

In consecutive training iterations, each mini-batch contains $K$ normal and $M$ anomalous video clips. The encoding and projection layers create visual representations, $v_{ni}$, and $v_{ai}$, $i \in \{1, \ldots, K+M\}$ from normal, and anomalous training clips, respectively. Each mini-batch contains $K(K-1)$ normal pairs and $KM$  anomalous pairs, which are input to the contrastive loss function. According to Equations \ref{eq:loss_func} and \ref{eq:total_loss}, the contrastive loss calculates the similarity between normal and anomalous pairs by calculating the dot product between the latent representations $(v_{ni}^T v_{nj}$ and $v_{ni}^T v_{am}$, scaled by a temperature parameter $\tau$ and exponential function. The numerator comprises of the similarity of normal pair and the denominator contains the similarity between normal pair and an aggregate of the similarities of all the anomalous pairs. Minimization of this objective function results in joint training of the encoding and projection layers to maximize the similarities between normal samples and minimize the similarities between normal and anomalous samples. However, the summation of similarities in anomalous pairs could be problematic in cases where anomalous samples are similar to normal samples, either due to nature of the data, mislabeling or noise. In such cases, the overall sum of the negative pairs could become larger, which is undesirable for the task and makes optimization harder. In order to overcome this issue, we propose to scale the sum of anomalous pairs s.t. the overall similarity of aggregated anomalous pairs remains smaller (modification \#1).
Scaling the summation of anomalous pairs with a very small number could lead to learning trivial representation. Therefore, we propose to use the average of summation of negative pairs (see Equation \ref{eq:loss_func}). Other possibilities for scaling could be explored by treating it as a hyper parameter.

\begin{equation}
\label{eq:loss_func}
    \mathcal{L}_{ij} = - log \frac{exp(v_{ni}^T v_{nj}/\tau)}{exp(v_{ni}^T v_{nj}/\tau) + \frac{1}{M} \sum_{m=1}^M exp(v_{ni}^T v_{am}/\tau)}
\end{equation}

\begin{equation}
\label{eq:total_loss}
    {L} = \frac{1}{K(K-1)} \sum_{i=1}^K \sum_{j=1}^K \mathbbm{1}_{j \neq i} \mathcal{L}_{ij}
\end{equation}

\noindent where $L$ is the loss function for each batch. 

After training with contrastive loss, 
projection layers are normally discarded in test phase \cite{chen2020simple,kopuklu2021driver, khosla2020supervised}. The rationale for this choice comes from the work of Chen et al. \cite{chen2020simple} in the context of self-supervised contrastive learning  that the latent representation, $h_i$, learned at the encoding stage will be more general and act as a pre-trained model for further classification task. However, we argue that for the supervised learning task, the representations learned at the projection layers ($v_i$) can be more useful by virtue of the contrastive loss learnt at the projection layers and 
aligns with the supervised learning task of keeping normal and anomalous classes far apart (modification \#2).

In the DAD dataset, the entire anomalous clips were given one label. A closer inspection suggested that these clips may not contain anomalous actions for the entire duration. Thus, two members of the research team manually re-labelled only the `anomalous' frames in those clips and discarded rest of the frames.
While manually re-labeling, we categorize a frame as normal driving behaviour if the participant has both the hands on the wheel and if the angle between the front facing direction and their current gaze is approximately smaller than $60$ degrees, as determined by visual inspection. Any frame that does not meet the above criteria is considered as anomalous driving behaviour. The normal driving clips were not re-labelled to keep variations of typical driving behaviours in those clips. 
This additional step is indeed specific to this dataset; however, the intent was to improve the quality of the training data. The labels in the test set remain unchanged for a fair comparison. We expect manual labeling to perform better because there will be less overlap with the normal driving in anomalous clips; thus, preventing the model from learning conflicting information. 

During the test phase, the jointly trained network is used for obtaining latent visual representation for each test sample. 
A template for the normal driving, ($v_n$) \cite{kopuklu2021driver}, can be calculated from the latent representations obtained at the projection head of all the normal clips:

\begin{equation}
\label{eq:vn}
    v_n = \frac{1}{N} \sum_{i=1}^N \frac{g_\beta (x_i)}{\norm{g_\beta (x_i)}_2}
\end{equation}

For a test clip, $t_i$, cosine similarity can be calculated between its latent representation at the projection head and the normal template as follows:

\begin{equation}
\label{eq:cosine_sim}
    s_i = v_n^T \frac{g_\beta (t_i)}{\norm{g_\beta (t_i)}_2}
\end{equation}

\noindent where $s_i$ is the similarity score that can be used for further analysis. The area under the curve (AUC) of the Receiver Operating Characteristic (ROC) and Precision-Recall (PR) curve are used as performance metrics. Due to the imbalance between normal and anomalous driving samples, AUC PR is a useful metric to highlight the relevance of a positive result given the baseline probabilities of a problem.

\section{Experimentation Details}

\subsection{Data Processing}

As the DAD dataset contains depth and IR modalities, video frames in each modality were converted to gray-scale (single-channel) images and normalized in the $[0,1]$ range. In the spatial domain, video frames were resized and center cropped to $112 \times 112$ pixels. In the temporal domain, non-overlapping windows of length $32$ were extracted and then down-sampled to $16$ frames.

During training, data augmentation in the form of random cropping, random rotation, salt and pepper noise, and random horizontal flipping (top-view only) were applied to consecutive frames of training samples \cite{kopuklu2021driver}. With the original labeling, 
$46950$ normal and $8600$ anomalous training windows are available in the DAD dataset.
As a result of manual labeling, the number of anomalous windows has been reduced to $7878$ because the normal windows that had been mistakenly identified as anomalous have been removed.
The number of normal training windows, however, remains the same. 
Accordingly, the imbalance ratio in this dataset with original and manual labeling is $5.45$ and $5.95$, respectively.


\subsection{Training Phase}


The encoder is a 3D ResNet-18 that generates a $512$-D tensor for each input video clip. The projection head is a $512$-$256$-$128$ feed-forward neural network whose input is the $512$-D output of the encoder. The output of the projection head is a normalized $128$-D tensor. 
The models were trained for $250$ epochs on a server with $64$ GB of RAM and an NVIDIA Tesla P100 PCIe $12$ GB GPU. Stochastic gradient descent was used with an initial learning rate of $0.01$ that was reduced by a factor of $10$ every $100$ epochs. The temperature parameter, $\tau$, was set to $0.1$. Each mini-batch contained $160$ windows in total, with $10$ normal driving windows and $150$ anomalous driving windows. All these settings were kept as same as Kopuklu et al. \cite{kopuklu2021driver}. The training samples in the DAD dataset were divided into two sets of validation set ($20\%$ of samples) and training set ($80\%$ of samples) with the same ratios of normal and anomalous samples. 
After every $10$ training epochs, the performance of the training models (the base encoder and projection head) was validated on the validation set in terms of AUC ROC, and the models (the base encoder and the projection head) achieving the highest AUC were saved as the best models (or replaced the best models so far). The saved best models after $250$ training epochs were considered as the final trained models and were used for testing. 
We present the results on using both the representations learned at $f_\theta$ and $g_\beta$ and show that latter works better in the supervised classification task.

\subsection{Test Phase}
\label{sec:testing_phase}
Kopuklu et al. \cite{kopuklu2021driver} tested their models on the original $32$ fps, whereas the models were trained on down-sampled $16$ fps video data. They mentioned a scoring method that computes score from a $16$ length window and slides it by one frame to give score to each middle frame. Since they gave score to each frame, they operated in the original fps and ignored that the training and testing models were incompatible and non-comparable. We identified this problem, created non-overlapping windows, down-sampled the testing data from $32$ fps to $16$ fps, and computed cosine similarity score for each of the non-overlapping windows w.r.t. to the normal template (see Equation \ref{eq:cosine_sim}) and calculated AUC ROC and PR from them. For the test phase, each batch contains $25$ windows, with each window containing $16$ frames. 

A separate model was trained for each modality and view, i.e., 
top and front view, IR and depth modality (four models). These models were tested for nine modality combinations -- four individual modalities and views (Top (D), Top (IR), Front (D), Front (IR)), two combinations for top and front for depth+IR (Top (DIR) and Front (DIR)), two combinations for depth and IR for top+front (Fusion (D) and Fusion (IR)), and one for combining all four top, front, IR and depth (Fusion (DIR)), where D and IR represents depth and infrared cameras. 
Different modalities and views were combined to verify if their fusion works better for detecting anomalous driving behaviours. In these cases, $f_\theta$ and $g_\beta$ representations are learned for each modality and/or view combination. During testing, corresponding windows are passed through $f_\theta$ and $g_\beta$ (from trained models). The average of these similarity scores 
is taken as a score for calculating AUC ROC and PR.
The code of our implementations is available at \url{https://github.com/abedicodes/SCL-DAD.}


\section{Results}

\begin{table*}
\centering
\begin{tabular}{|p{47pt}|p{48pt}|p{41pt}||p{41pt}|p{41pt}|||p{41pt}|p{41pt}||p{41pt}|p{41pt}|} \hline
\multirow{3}{*}{\textbf{Modality}}
  & \multicolumn{4}{c|||}{\textbf{Original Loss}} & \multicolumn{4}{c|}{\textbf{Proposed Loss}} \\ \cline{2-9}
  & \multicolumn{2}{c||}{\textbf{No Projection Head}} & \multicolumn{2}{c|||}{\textbf{Projection Head}} & \multicolumn{2}{c||}{\textbf{No Projection Head}} & \multicolumn{2}{c|}{\textbf{Projection Head}} \\ \cline{2-9}
  & \textbf{Original labeling} & \textbf{Manual labeling} & \textbf{Original labeling} & \textbf{Manual labeling} & \textbf{Original labeling}  & \textbf{Manual labeling} & \textbf{Original labeling} & \textbf{Manual labeling} \\ \hline
Top (D) & 0.8713, 0.9128 & 0.8573, & 0.8876 & 0.8604 & 0.8983 & 0.9080 & 0.8722 & \cellcolor{gray!50}0.9202 \\ \hline
Top (IR) & 0.8456, 0.8804 & 0.8423 & 0.8406 & 0.8642 & 0.8118 & 0.8769 & 0.7952 & \cellcolor{gray!50}0.8913 \\ \hline
Top (DIR) & 0.8663, 0.9166 & 0.9007 & 0.8881 & 0.9021 & 0.8919 & 0.8969 & 0.8592 & \cellcolor{gray!50}0.9273 \\ \hline
Front (D) & 0.8503, 0.8996 & \cellcolor{gray!50}0.9156 & 0.8577 & 0.9064 & 0.8759 & 0.8773 & 0.8461 & 0.8851 \\ \hline
Front (IR) & 0.8367, 0.8695 & 0.8532 & 0.8324 & 0.8823 & 0.8854 & 0.8782 & 0.8448 & \cellcolor{gray!50}0.9113 \\ \hline
Front (DIR) & 0.8818, 0.9196 & \cellcolor{gray!50}0.9302 & 0.8998 & 0.8945 & 0.8969 & 0.9250 & 0.8722 & 0.9268 \\ \hline
Fusion (D) & \cellcolor{gray!50}0.9229, 0.9609 & 0.9514 & 0.9082 & 0.9042 & 0.9449 & 0.9451 & 0.9154 & 0.9517 \\ \hline
Fusion (IR) & 0.8946, 0.9321 & 0.8937 & 0.8941 & 0.9289 & 0.9240 & 0.9337 & 0.8932 & \cellcolor{gray!50}0.9552 \\ \hline
Fusion (DIR) & 0.9342, 0.9655 & 0.9536 & 0.9248 & 0.9451 & 0.9599 & 0.9619 & 0.9366 & \cellcolor{gray!50}0.9738 \\ \hline
\end{tabular}
\caption{AUC of ROC curve results corresponding to different combinations of losses (original or proposed), projection head (present or absent), and labeling (original or manual). The modality abbreviations used are: Depth (D), Infrared Radiation (IR), Depth and IR (DIR).
Each cell in the second column contains two numbers; the first being the AUC calculated after modifying the problem in \cite{kopuklu2021driver} (as described in section \ref{sec:testing_phase}), and the second number is the original AUC reported in \cite{kopuklu2021driver}.}

\label{tab:roc_our_method}
\end{table*}

\begin{table*}
\centering
\begin{tabular}{|p{47pt}|p{48pt}|p{41pt}||p{41pt}|p{41pt}|||p{41pt}|p{41pt}||p{41pt}|p{41pt}|} \hline
\multirow{3}{*}{\textbf{Modality}}
  & \multicolumn{4}{c|||}{\textbf{Original Loss}} & \multicolumn{4}{c|}{\textbf{Proposed Loss}} \\ \cline{2-9}
  & \multicolumn{2}{c||}{\textbf{No Projection Head}} & \multicolumn{2}{c|||}{\textbf{Projection Head}} & \multicolumn{2}{c||}{\textbf{No Projection Head}} & \multicolumn{2}{c|}{\textbf{Projection Head}} \\ \cline{2-9}
  & \textbf{Original labeling} & \textbf{Manual labeling} & \textbf{Original labeling} & \textbf{Manual labeling} & \textbf{Original labeling}  & \textbf{Manual labeling} & \textbf{Original labeling} & \textbf{Manual labeling} \\ \hline
Top (D) & 0.8925 & 0.8734 & 0.9097 & 0.8994 & \cellcolor{gray!50}0.9253 & 0.9152 & 0.8979 & 0.9243 \\ \hline
Top (IR) & 0.8764 & 0.9152 & 0.8985 & 0.9062 & \cellcolor{gray!50}0.9226 & 0.9221 & 0.8512 & 0.9022 \\ \hline
Top (DIR) & 0.8910 & 0.9154 & 0.9005 & 0.9019 & 0.9162 & 0.9115 & 0.8947 & \cellcolor{gray!50}0.9182 \\ \hline
Front (D) & 0.8900 & 0.9127 & 0.8922 & \cellcolor{gray!50}0.9373 & 0.9274 & 0.9257 & 0.8952 & 0.9176 \\ \hline
Front (IR) & 0.8997 & 0.9016 & 0.9090 & 0.9457 & 0.9232 & \cellcolor{gray!50}0.9542 & 0.8931 & 0.9024 \\ \hline
Front (DIR) & 0.9002 & 0.9218 & 0.9123 & 0.9304 & 0.9358 & \cellcolor{gray!50}0.9607 & 0.9138 & 0.9284 \\ \hline
Fusion (D) & 0.9321 & 0.9400 & 0.9312 & 0.9502 & 0.9498 & \cellcolor{gray!50}0.9611 & 0.9439 & 0.9497 \\ \hline
Fusion (IR) & 0.9020 & 0.9547 & 0.9442 & 0.9518 & 0.9254 & \cellcolor{gray!50}0.9728 & 0.9209 & 0.9334 \\ \hline
Fusion (DIR) & 0.9618 & 0.9578 & 0.9695 & 0.9544 & \cellcolor{gray!50}0.9772 & 0.9721 & 0.9427 & 0.9591 \\ \hline
\end{tabular}
\caption{AUC of PR curve results corresponding to different combinations of losses (original or proposed), projection head (present or absent), and labeling (original or manual). The modality abbreviations used are: Depth (D), Infrared Radiation (IR), Depth and IR (DIR).}
\label{tab:pr_our_method}
\end{table*}

\begin{table*}
\centering
\begin{tabular}{|p{50pt}|p{42pt}|p{42pt}|p{42pt}|p{42pt}|p{40pt}|p{40pt}|p{40pt}|p{40pt}|} \hline
  & \textbf{OL-NP-NL}  & \textbf{OL-NP-ML} & \textbf{OL-PH-NL} & \textbf{OL-PH-ML} & \textbf{PL-NP-NL}  & \textbf{PL-NP-ML} & \textbf{PL-PH-NL} & \textbf{PL-PH-ML} \\ \hline
\textbf{OL-NP-NL} & n/a & 0.743 & 1.000 & 0.743 & 0.743 & 0.079 & 1.000 & \cellcolor{gray!50}{\bf0.001} \\ \hline
\textbf{OL-NP-ML} & 0.743 & n/a & 0.743 & 1.000 & 1.000 & 1.000 & 0.150 & 0.210 \\ \hline
\textbf{OL-PH-NL} & 1.000 & 0.743 & n/a & 0.743 & 0.743 & 0.079 & 1.000 & \cellcolor{gray!50}{\bf0.001} \\ \hline
\textbf{OL-PH-ML} & 0.743 & 1.000 & 0.743 & n/a & 1.000 & 1.000 & 0.210 & 0.150 \\ \hline
\textbf{PL-NP-NL} & 0.743 & 1.000 & 0.743 & 1.000 & n/a & 1.000 & 0.210 & 0.150 \\ \hline
\textbf{PL-NP-ML} & 0.079 & 1.000 & 0.079 & 1.000 & 1.000 & n/a & \cellcolor{gray!50}{\bf0.008} & 1.000 \\ \hline
\textbf{PL-PH-NL} & 1.000 & 0.150 & 1.000 & 0.210 & 0.210 & \cellcolor{gray!50}{\bf0.008} & n/a & \cellcolor{gray!50}{\bf0.000} \\ \hline
\textbf{PL-PH-ML} & \cellcolor{gray!50}{\bf0.001} & 0.210 & \cellcolor{gray!50}{\bf0.001} & 0.150 & 0.150 & 1.000 & \cellcolor{gray!50}{\bf0.000} & n/a \\ \hline
\end{tabular}
\caption{Friedman's post-hoc test with Bergmann and Hommel’s correction for different combinations of loss, projection head, and labeling across all the camera modalities. The test was performed on the AUC of ROC curve results. The abbreviations are: OL - Original Loss, NP - No Projection Head, NL - no labeling (use original video labels as is), PL - Proposed Loss, PH - Projection Head, ML - Manual labeling.}
\label{tab:stats_roc_our_method}
\end{table*}

\begin{table*}
\centering
\begin{tabular}{|p{50pt}|p{42pt}|p{42pt}|p{42pt}|p{42pt}|p{40pt}|p{40pt}|p{40pt}|p{40pt}|} \hline
  & \textbf{OL-NP-NL}  & \textbf{OL-NP-ML} & \textbf{OL-PH-NL} & \textbf{OL-PH-ML} & \textbf{PL-NP-NL}  & \textbf{PL-NP-ML} & \textbf{PL-PH-NL} & \textbf{PL-PH-ML} \\ \hline
\textbf{OL-NP-NL} & n/a & 0.652 & 1.000 & \cellcolor{gray!50}{\bf0.033} & \cellcolor{gray!50}{\bf0.001} & \cellcolor{gray!50}{\bf0.000} & 1.000 & 0.068 \\ \hline
\textbf{OL-NP-ML} & 0.652 & n/a & 1.000 & 1.000 & 0.242 & 0.161 & 0.917 & 1.000 \\ \hline
\textbf{OL-PH-NL} & 1.000 & 1.000 & n/a & 0.815 & 0.078 & \cellcolor{gray!50}{\bf0.046} & 1.000 & 1.000 \\ \hline
\textbf{OL-PH-ML} & \cellcolor{gray!50}{\bf0.033} & 1.000 & 0.815 & n/a & 1.000 & 1.000 & 0.068 & 1.000 \\ \hline
\textbf{PL-NP-NL} & \cellcolor{gray!50}{\bf0.001} & 0.242 & 0.078 & 1.000 & n/a & 1.000 & \cellcolor{gray!50}{\bf0.002} & 1.000 \\ \hline
\textbf{PL-NP-ML} & \cellcolor{gray!50}{\bf0.000} & 0.161 & \cellcolor{gray!50}{\bf0.046} & 1.000 & 1.000 & n/a & \cellcolor{gray!50}{\bf0.001} & 0.917 \\ \hline
\textbf{PL-PH-NL} & 1.000 & 0.917 & 1.000 & 0.068 & \cellcolor{gray!50}{\bf0.002} & \cellcolor{gray!50}{\bf0.001} & n/a & 0.161 \\ \hline
\textbf{PL-PH-ML} & 0.068 & 1.000 & 1.000 & 1.000 & 1.000 & 0.917 & 0.161 & n/a \\ \hline
\end{tabular}
\caption{Friedman's post-hoc test with Bergmann and Hommel’s correction for different combinations of loss, projection head, and labeling across all the camera modalities. The test was performed on the AUC of PR curve results. The abbreviations are: OL - Original Loss, NP - No Projection Head, NL - no labeling (use original video labels as is), ML - Proposed Loss, PH - Projection Head, ML - Manual labeling.}
\label{tab:stats_pr_our_method}
\end{table*}

Tables \ref{tab:roc_our_method} and \ref{tab:pr_our_method} show the AUC ROC and AUC PR results. The rows in the tables show the camera modality and the columns show different combinations of loss (original and proposed), projection head (absent or present), and labeling (original or manual). The highlighted cells indicate the highest AUC for that modality for a given combination of loss, projection head, and labeling. 
The baseline results for each modality corresponds to the combination of original loss, no projection head, and original labeling.
Each cell in the second column of Tables \ref{tab:roc_our_method} contains two numbers, the first number is the AUC we calculated after modifying the problem in \cite{kopuklu2021driver} (as described in section \ref{sec:testing_phase}), and the second number is the original AUC reported in \cite{kopuklu2021driver}.
We observe that in terms of AUC ROC, out of nine modalities, the proposed loss with the presence of projection head and manual labeling performed better in
six of them (improved performance from $4.23\%$ for Fusion(DIR) to $8.91\%$ for Front(IR)).
In the rows where the last column's result is the best, the percentage increase is calculated by comparing this number with the baseline (the first number in the first column).
In two cases, original loss with no projection head and manual labeling performed better (Front (D) and Front (DIR)). However, 
eight of the nine best performing models had manual labeling, further highlighting the role of good quality of labels in supervised classification problems. 
Some other findings are as follows:
\begin{itemize}
    \item Fusion of camera modalities (D and IR) for each top and front view improved the performance in comparison to individual camera and modality type irrespective of loss type and its various combinations.
    
    \item Major performance improvement was achieved when top and front views were fused for each of the depth and IR cameras -- Fusion (D) (AUC=
    $0.9609$)
    and Fusion (IR) (AUC=$0.9552$).
    Combining different views can help in capturing complementary information which improved the performance.
    
    \item The best performance was achieved when both the camera modalities and views were combined (AUC=$0.9738$). Thus, 
    apart from 
    top and front views, combining depth and IR cameras further boost the performance in detecting anomalous driving behaviours.
\end{itemize}

In terms of AUC PR (see Table \ref{tab:pr_our_method}), eight out of nine highest numbers are for proposed loss and distributed across different combinations of projection head and manual labeling. This further ascertains that modifying the loss function is important in a supervised contrastive learning task, and adding projection head and manual labeling further improves the performance.

\subsection{Statistical Testing}
To understand the overall performance improvement of the
proposed method,
we performed the Friedman’s post-hoc test with Bergmann and Hommel’s correction \cite{calvo2016scmamp, scmamp, statspackage}. This test computes the p-value for each pair of algorithms corrected for multiple testing using Bergman and Hommel’s correction. Table \ref{tab:stats_roc_our_method} shows the (AUC ROC) comparison between different combinations of loss, projection head, and labeling across nine modalities. The significant differences (p-value $< 0.05$) are shown in bold and highlighted cells.
From Table \ref{tab:stats_roc_our_method}, we infer the following:
\begin{itemize}
    \item The performance of the proposed loss with the presence of the projection head and manual labeling is statistically different than original loss and original labeling (irrespective of projection head) and proposed loss with projection head and original labeling. 
    \item The proposed loss with manual labeling (irrespective of projection head) is statistically different than proposed loss with projection head and original labeling. 
    
    \item Within original loss, each combination performs statistically similar.
\end{itemize}

Table \ref{tab:stats_pr_our_method} shows the (AUC PR) comparison between different combinations of loss, projection head, and labeling across nine modalities, we infer the following:
\begin{itemize}
    \item The proposed loss with no projection head (irrespective of labeling) is statistically different than the proposed loss with projection head and original labeling and the baseline (original loss with no projection head and original labeling).
    
    \item The proposed loss with no projection and manual labeling is statistically different than original loss with projection head, and original labeling.
    
    \item Original loss with projection head and manual labeling is statistically different than original loss with no projection head and original labeling (baseline) and statistically similar to other original loss combinations. 
\end{itemize}

Therefore, combining the results from Tables \ref{tab:stats_roc_our_method} and \ref{tab:stats_pr_our_method}, we conclude that within a supervised classification framework, the proposed contrastive loss with manual labeling is certainly superior and adding the projection head can also help in detecting seen and unseen anomalous driving behaviours.

\section{Conclusions} 
We presented a new supervised contrastive loss function
along with considerations for including projection head and refining labels to improve the detection of anomalous driving behaviours from videos. We support our results through rigorous statistical testing. We found that combining video clips from different views improves the performance drastically, with the best results achieved by combining top and frontal views and depth and IR cameras. 

\bibliographystyle{IEEEtran}
\bibliography{myrefs}

\end{document}